\documentclass[a4paper,twoside]{article}

\usepackage{epsfig}
\usepackage{subcaption}
\usepackage{calc}
\usepackage{amssymb}
\usepackage{amstext}
\usepackage{amsmath}
\usepackage{amsthm}
\usepackage{multicol}

\usepackage{multirow}
\usepackage{booktabs}
\usepackage{makecell}
\usepackage{arydshln}
\usepackage[usenames,dvipsnames]{xcolor}
\usepackage{float}
\usepackage{graphics}

\usepackage{pslatex}
\usepackage{apalike}
\usepackage{algorithm2e}
\usepackage[bottom]{footmisc}
\usepackage{SCITEPRESS}    

\begin{document}

\title{Improved MambdaBDA Framework for \\ Robust Building Damage Assessment Across Disaster Domains}

\author{\authorname{Alp Eren Gençoğlu\sup{1} and Hazım Kemal Ekenel\sup{1,2}}
\affiliation{\sup{1}Department of Computer Engineering, Istanbul Technical University, Türkiye}
\affiliation{\sup{2}Division of Engineering, New York University Abu Dhabi, UAE}
\email{\{gencoglu17, ekenel\}@itu.edu.tr}
}

\keywords{Remote Sensing, Building Damage Assessment, Change Detection, Focal Loss, Attention Gates}

\abstract{Rapid and reliable post-disaster building damage assessment from satellite imagery is hindered by severe class imbalance, background clutter, and domain shift across disaster types and geographies. In this work, we address these problems and explore ways to improve the MambaBDA, the building damage assessment network of ChangeMamba architecture, one of the most successful building damage assessment models. The approach enhances the MambaBDA with three modular components: (i) Focal Loss to mitigate class imbalance in four-level damage classification, (ii) lightweight Attention Gates to suppress irrelevant context in both localization and damage heads, and (iii) a compact Alignment Module to spatially warp pre-event features toward post-event content before decoding. We experiment on multiple satellite imagery datasets, including xBD, Pakistan Flooding, Turkey Earthquake, and Ida Hurricane, and conduct both in-domain and cross-dataset tests. The proposed modular enhancements yield consistent improvements over the baseline model, ranging from $0.8\%$ to $5\%$ performance gains in in-domain tests and even more significant performance gains, up to $27\%$, on unseen disasters. This indicates that the proposed enhancements are especially beneficial for the generalization capability of the system.}

\onecolumn \maketitle \normalsize \setcounter{footnote}{0} \vfill

\section{\uppercase{Introduction}}
\label{sec:introduction}

Building Damage Assessment (BDA) is the task of identifying and classifying physical damage on any type of building, after a catastrophic event or a disaster, such as an earthquake, wildfire, flood, or hurricane. This process is carried out by obtaining the pre-disaster image and post-disaster image of the same scene, and distinguishing the changes on the buildings between the two time frames. Fundamentally, it is a localization and change detection (CD) task. The output of this process yields a BDA map which highlights buildings and damage levels for the specific disaster region and helps authorities to take action. Timely and accurate BDA maps carry a great importance for search-and-rescue and engineering teams guidance, loss estimation, and economic impact assessment for the government and insurance agencies, and long-term recovery planning.

With the rapid development of Deep Learning~(DL) methods and big data, approaches for CD task have also been improved in recent years and achieved decent success rates. In this field, specifically for the BDA task, remote sensing, particularly high-resolution satellite imagery has been a greatly valuable tool for a long time. The increase of the number and quality of the disaster-based satellite imagery datasets also accelerated the improvements. The momentum increased with large-scale datasets and benchmarks, such as the xView \cite{xView1_lam2018xviewobjectscontextoverhead} challenge and xBD dataset \cite{xBD_xView2_gupta2019xbddatasetassessingbuilding}. 

The deep neural network architectures proposed for the task have also been evolving over the years. Until recently, convolutional neural network (CNN) based models were the most widely used architecture types \cite{CNN_Survey1_Deep_Convolutional_Neural_Networks_for_Image_Classification_A_Comprehensive_Review,CNN_Survey2_MedicalImageAnalysis_Litjens_2017}. High speed and great local recognition capabilities made CNNs very successful, however they lacked accuracy in long-range pixel dependencies.  After the self-attention and transformers are integrated for visual tasks, Vision Transformer \cite{vit_dosovitskiy2021imageworth16x16words} based models also became increasingly more popular. The quadratic complexity of the self-attention, however, makes it computationally very expensive. As a less complex and highly efficient alternative, Mamba architecture \cite{Mamba_gu2024mambalineartimesequencemodeling}, which is a state space model (SSM), and VMamba \cite{VMamba_liu2024vmambavisualstatespace} (Visual SSM) are introduced. Finally, ChangeMamba architecture \cite{ChangeMamba_Chen_2024} is a state-of-the-art CD architecture, with a network specifically designed for the BDA task, MambaBDA. 

Despite the progress, current state of the art still faces some challenges that are posed by the BDA task. Firstly, most of the available large-scale datasets have class imbalance problem. For example, they have much more "not damaged" samples than "major damage" samples. This makes it more difficult for the model to learn without bias. Secondly, background variance, lighting, and shadows, etc. can also make it harder to distinguish between building and background, causing false positives. Lastly, even though most of the datasets are said to be registered (pre-disaster and post-disaster images are aligned), there can be some small misalignment and pixel shifts due to images being taken in different times and with different satellite angles. These problems can make it more difficult for the model to perform better and even more difficult to generalize to unseen data.

In this study, we investigate whether implementing lightweight modular enhancements can provide an improvement to the  MambaBDA model's performance, both for in-domain and cross-dataset (on unseen data) tests. In summary, the following are the main contributions of this paper:
\begin{enumerate}
    \item We include focal loss, alongside the originally utilized cross-entropy and Lovász-Softmax losses, to mitigate class imbalance problem.
    \item We integrate Attention Gate mechanisms into decoder modules to suppress irrelevant features from the encoder and obtain better segmentation results. 
    \item We introduce a custom lightweight Alignment Module to dynamically compensate for minor spatial misalignments between pre-disaster and post-disaster images.
    \item We provide the analysis of the in-domain and cross-dataset tests carried out using multiple variations of the model and multiple large-scale damage assessment datasets. 
\end{enumerate}

The remainder of this paper is structured as follows: Section \ref{sec:related_work} reviews related work, including datasets and deep learning methods in the field. Section \ref{sec:methodology} details the methodology, including MambaBDA baseline and our focal loss, attention gate, alignment enhancements. Section \ref{sec:experimental_setup} describes datasets, evaluation metrics, training details, and experiment groups. Section \ref{sec:results} delves into the experimental results. Finally, Section \ref{sec:conclusion} concludes the paper.


\section{\uppercase{Related Work}}
\label{sec:related_work}

In this section, we focus on the datasets used in BDA tasks and deep learning based methods developed for the CD tasks.

\subsection{BDA Datasets \& Benchmarks}
For the BDA task, a large, pre/post-event paired, high-resolution dataset is required. The xView 2 \cite{xBD_xView2_gupta2019xbddatasetassessingbuilding} Challenge extended the original xView object-detection dataset \cite{xView1_lam2018xviewobjectscontextoverhead}, and introduced the xBD dataset \cite{xBD_xView2_gupta2019xbddatasetassessingbuilding}. The xBD dataset consists of satellite image pairs of multiple different disasters and regions. Since then, it has become the practical standard for BDA benchmarking thanks to its large-scale size, high resolution and diversity. Many datasets after it also used similar system for labeling the damages. In xBD dataset, there are four damage classes; "No-damage", "Minor", "Major", and "Destroyed". The labels are provided in a JSON file, annotating the building borders using polygon coordinates and categorizing them with one of the four damage classes. Data co-registration, aligning the pre-disaster and post-disaster images spatially, is also a very important concept for BDA datasets, which xBD provides.

The Extensible Building Damage (EBD) dataset \cite{EBD_Dataset_Wang2024} is another, and more recent large-scale BDA dataset, or rather collection of datasets. It is created by gathering 12 disaster events, which are not included in the xBD dataset, in different countries. The dataset includes a variety of disasters such as earthquake, hurricane, and flooding. 

There are also individual small scale BDA datasets, such as mwBTFreddy \cite{mwBTFreddy_chapuma2025mwbtfreddydatasetflashflood} and Ida-BD \cite{Ida-BD_lee2022ida} which are not very suitable to use for directly training complex deep learning models individually, due to their insufficient amount of data. 

\subsection{Deep Learning in Change Detection}

The field of remote sensing change detection has evolved rapidly over the years. Earlier state of the art models mostly relied on Convolution Neural Networks (CNNs). Since CNNs use filters and mostly operate on local spatial correlations, they have high level capabilities of especially low and mid-level feature extraction. They are also computationally effective and do not require too many parameters and high memory size. The U-Net \cite{U-NET_ronneberger2015unetconvolutionalnetworksbiomedical} is one of the significant architectures of its time, which was created for biomedical image segmentation and set a strong foundation for pixel-level classification tasks.

After a few years, an improved version of U-Net was introduced, which especially upgrades spatial focusing, by a new approach named Attention Gates \cite{attention-gate-unet_oktay2018attentionunetlearninglook}. Attention Gate (AG) is a module that learns to suppress feature activations in the irrelevant parts of the image and strengthens the activations in task-relevant regions by creating masks and modulating skip connections. 

One of the most significant limitations of CNNs is their low success rate on capturing long range context, stemming from the inherent filter based working principle of CNNs. To fill this gap, Transformer \cite{transformer_vaswani2023attentionneed} based architectures were introduced. The Transformer model was initially proposed for the natural language processing (NLP) domain, but shortly after it was adapted into computer vision domain, one of the prominent representatives being Vision Transformer (ViT) architecture \cite{vit_dosovitskiy2021imageworth16x16words}. Transformers' main mechanism is self-attention, which is used for establishing a relation between each token in the given input (each pixel for images). This way, the model is able to capture long range relations and global context. This enables the Transformer based models to have higher scores on certain benchmarks, but at a cost of being more computationally expensive, because of the $O(N^2)$ operations required for self-attention mechanism, where $N$ is the number of tokens/pixels. A model that stands out for utilizing Transformers for CD tasks is Bitemporal Image Transformer (BIT) \cite{BIT_Chen_2022} which creates "semantic tokens" to represent the image in fewer tokens, modeling the context in an efficient way, and utilizes cross-temporal attention to grab the context between pre/post disaster images. After that, ChangeFormer \cite{ChangeFormer_bandara2022transformerbasedsiamesenetworkchange} proposed a siamese transformer network, reporting gains on standard CD benchmarks.

Although Transformer based architectures are successful, their memory and computation requirements scale quadratically, and they require very large amounts of data since they lack the inductive bias capabilities of CNNs \cite{ScratchFormer_noman2023remotesensingchangedetection}. To fix this problem, Structured State Space Sequence model (S4) was introduced to provide prior aproaches' advantages, combining properties of Recurrent Neural Networks (RNNs) and CNNs, while being more efficient \cite{S4_gu2022efficientlymodelinglongsequences}. The Mamba architecture then introduced selective scan mechanism into the S4 model, which is named as an S6 model, enabling linear-time, $O(N)$, sequence modeling \cite{Mamba_gu2024mambalineartimesequencemodeling}. This makes it a very powerful architecture for long sequence modeling, having both local efficiency and global context capturing capabilities. For the vision-based tasks, VMamba introduced the Visual State Space (VSS) module that integrates a 2D Selective Scan (SS2D) mechanism to propose a vision backbone \cite{VMamba_liu2024vmambavisualstatespace}.

ChangeMamba \cite{ChangeMamba_Chen_2024} is introduced as a CD task solution, employing VMamba's VSS modules in its backbone. ChangeMamba proposes three networks; a Binary CD model, a Semantic CD model, and a BDA model. The latter, MambaBDA, model is utilized in this study. MambaBDA consists of the encoder network, which is a weight-sharing siamese network based on VMamba, and the decoder network with two heads, one for building localization and one for damage classification (pixel-wise).

Despite having a high performance, MambaBDA model encounters difficulties due to the nature of the BDA task and datasets. In this study, we adopt MambaBDA as a baseline and explore the effects of modular enhancements to improve the performance. The employed modular enhancements are; focal loss for class imbalance, attention gates for regional suppression, and alignment module for more precise feature aligning.


\section{\uppercase{Methodology}}
\label{sec:methodology}
In this section, we first describe the baseline model utilized in this work, then explain our three modular enhancements; focal loss, attention gates, and alignment module. 

\subsection{Baseline Model}
\label{sec:baseline_model}
\begin{figure}[!h]
  \centering
   \includegraphics[width=\linewidth]{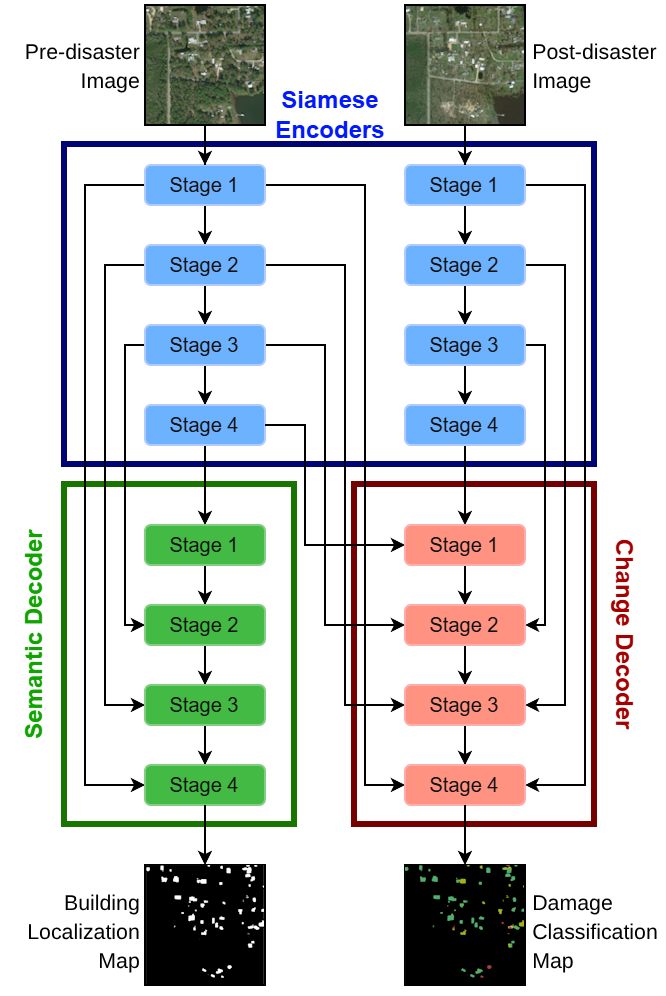}
  \caption{Network structure of MambaBDA, adapted from \cite{ChangeMamba_Chen_2024}.}
  \label{fig:MambaBDA_overall_structure}
 \end{figure}

As the baseline model, ChangeMamba's BDA architecture, MambaBDA, is selected as it is an efficient model that yields state-of-the-art performance. The architecture comes in three variations; tiny, small, and base. Since the most successful variant is the base one according to the original results \cite{ChangeMamba_Chen_2024}, MambaBDA-Base is selected. We refer to it \textbf{MambaBDA} hereafter. MambaBDA consists of two main blocks, encoder and decoder, where encoder is a weight-sharing siamese network for extracting pre-disaster and post-disaster image features. The \textbf{encoder} consists of four blocks each of which first downsamples the data and then models contextual information using a visual state space (VSS) block. This way, the encoder produces multi-level features for the decoders. The decoders on the other hand, consist of two separate streams; one head for building localization predictions, semantic decoder, and one head for damage classification predictions, change decoder, both of which also consisting of four stages. The \textbf{semantic decoder} fundamentally restores the building localization maps step by step, with the corresponding feature from pre-disaster encoder. This operation is done by first putting the feature through a VSS block, then upsampling it, then fusing the result with the feature from the lower level features. The \textbf{change decoder} utilizes features from both pre-disaster and post-disaster encoders. These features are fed into spatio-temporal state space (STSS) block which rearranges the data and puts through the VSS blocks. Then the result is fused with the previous stage's output and upsampled. Finally, the output of the change decoder yields a damage classification map. An overall structure of the network is demonstrated in Figure \ref{fig:MambaBDA_overall_structure}.

\subsection{Enhanced Models}
\label{sec:our_enhancements}

\subsubsection{Adopting Focal Loss}
\label{sec:focal_loss}
In order to mitigate the heavy skew towards the "no damage" class and the general imbalance problem between other classes, the damage head is trained with the focal loss \cite{Focal_Loss_lin2018focallossdenseobject}. Focal loss is similar to cross-entropy (CE) loss, but it introduces two extra parameters: 
\begin{itemize}
    \item $\gamma$ is the focusing parameter, used to make the model focus more on harder samples. Value is used as $>1$. This helps for extreme uneven class distributions, such as xBD having over 10 times more "no damage" pixels than "destroyed" pixels.
    \item $\alpha$ is the class weighting factor list, having a separate value for each possible class. The values can be determined by inverse class frequencies or can be fine-tuned depending on the model's outputs. This helps balancing model's confusion.
\end{itemize}
According to this, the focal loss is calculated as follows:
\begin{equation}
    \mathcal{L}_{\text{focal}} = - \frac{1}{N}\sum_{i=1}^{N} \sum_{c=1}^{C} \alpha_c\, y_{ic}\,(1-p_{ic})^{\gamma}\,\log(p_{ic}),
\end{equation}
where $N$ is the total number of pixels, $C$ is the number of classes, which is 4, $p_{ic}$ is the probability that pixel $i$ belongs to class $c$, and $y_{ic}$ is one-hot label, meaning $1$ if pixel $i$ belongs to class $c$, otherwise $0$. In our experiments, we found that $\alpha = [0.6, 1.6, 1.1, 1.1]$ and $\gamma = 1.5$ best suits for our specific use case for xBD dataset. For $\alpha$ parameter selection, we first started with inverse frequencies and then fine-tuned with model confusion.

MambaBDA is originally trained with CE and Lovász-Softmax losses. Since Lovász-Softmax \cite{Lovasz_Softmax_berman2018lovaszsoftmaxlosstractablesurrogate} serves a different purpose, optimizing Intersection over Union (IoU), we kept that as is, and for CE loss, we split it into a combination of CE and focal loss, since using only focal loss yielded unstable results. In summary, our final loss for the damage head is a combination of CE, Focal and Lovász-Softmax losses. Our final loss for the building head is left as the original implementation, not using focal loss.

\subsubsection{Attention Gate Integration}
\label{sec:attention_gate}
Attention Gates (AGs) \cite{attention-gate-unet_oktay2018attentionunetlearninglook} serve as a mechanism to filter out background clutter and emphasize relevant features in the decoder stream. Skip features often carry irrelevant information such as shadows, roads, bodies of water. To mitigate this problem and have the model make more precise predictions, we integrate lightweight AGs at each skip connection between the decoder stages. In total there are three AGs per decoder, enabling the model to focus on change-relevant regions dynamically. The basic form of an AG can be expressed by the following formula:
\begin{equation}
    \alpha_{attn} = \sigma_2( \psi( \sigma_1(W_xx + W_gg + b_g) ) + b_{\psi} ), 
\end{equation}
where $\sigma_1$ and $\sigma_2$ are activation functions and usually ReLU and sigmoid activations are used, respectively. Before the operations, the $g$ is upsampled to the shape of $x$. Since the result, $\alpha_{attn} \in [0,1]$, it is used as a mask and the final gated output is obtained by elementwise multiplication as follows:
\begin{equation}
    \hat{x} = \alpha_{attn} \odot x.
\end{equation}

We also included a normalization layer before the activations to make the training more stable. Since we deal with mostly small batch sizes, which is common in high resolution CD tasks, we integrated a group normalization (GN) \cite{GroupNorm_wu2018groupnormalization} layer instead of a batch normalization (BN) \cite{BatchNorm_ioffe2015batchnormalizationacceleratingdeep} layer. GN divides the channels into a number of groups and then calculates mean and standard-deviation separately over each group, making it more robust than BN especially for small batch sizes. For the gating part we also modified the formula to prevent complete suppression of the features, allowing a minimum 50\% signal retention. This helps maintaining gradient flow even when $\alpha_{attn} \approx 0$, and works similar to residual connections by always preserving some of the information. As a result, our implementation of AGs can be expressed by the following:
\begin{equation}
    \alpha_{attn} = \sigma_2( \psi( \sigma_1(\text{GN}(W_xx) + \text{GN}(W_gg + b_g)) ) + b_{\psi} ) ,
\end{equation}
\begin{equation}
    \hat{x} = (0.5 + 0.5\alpha_{attn}) \odot x,
\end{equation}
where $W_x$, $W_g$ and $\psi$ are used as $1 \times 1$ convolutional layers, $\sigma_1$ is ReLU function and $\sigma_2$ is sigmoid function. 

The AG modules are implemented separately for the building decoder and damage decoder. Each decoder has three AGs in the connection points of the four stages. The AG modules for a decoder can be enabled or disabled in our modular implementation, and can be tested separately.

\subsubsection{Custom Alignment Module}
\label{sec:alignment_module}
Small spatial misalignments between pairs of pre-disaster and post-disaster inputs can affect the model learning negatively. In order to address this problem, we introduce a shallow alignment module that learns to warp pre-features to post-features. 

The alignment module is a simple convolutional network, that applies $3 \times 3$ convolution three times with ReLU activation in-between. The module is placed between encoding and decoding phases, and it operates on the encoded features separately, stage by stage. The inputs of the network are the corresponding stage's pre-features and post-features. These two feature pairs are concatenated to obtain a single input object. Then it is passed through the convolutional networks. Finally, an output with the same height and width with 2 channels is obtained. One channel corresponds to horizontal shift and the other vertical.

Given the features $(f^{\text{pre}}, f^{\text{post}}) \in \mathbb{R}^{h \times w \times c}$, the module predicts an offset (flow) map in 2D: $\Delta \in \mathbb{R}^{h \times w \times 2}$. Then this flow is used to warp the pre-features to align with the post-features.


\section{\uppercase{Experimental Setup}}
\label{sec:experimental_setup}
In this section, we specify the experimental protocol. Section \ref{sec:datasets} details the properties of the used datasets and preprocessing. Sections \ref{sec:evaluation_metrics} and \ref{sec:training_details} describes the evaluation metrics and reports training settings, respectively. Section \ref{sec:experiment_groups} explains the experiment groups for in-domain and cross-dataset testing.

\subsection{Datasets}
\label{sec:datasets}
\begin{figure}
    \centering
    \includegraphics[width=\linewidth]{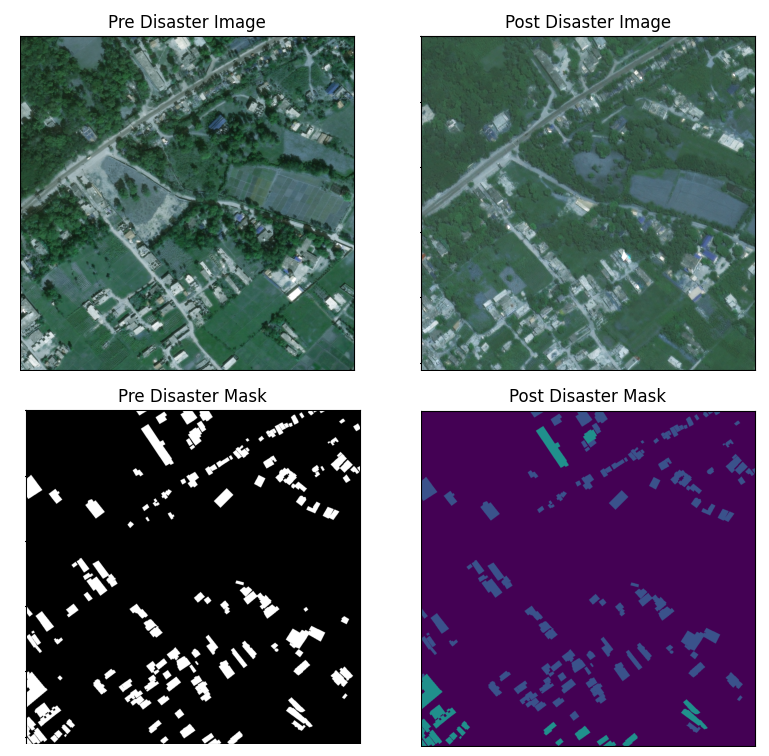}
    \caption{Sample data from xBD consisting of image pairs and mask pairs.}
    \label{fig:sample_xbd_data}
\end{figure}

In order to evaluate our methods, we make use of pre/post-event paired satellite imagery datasets that annotates both the building borders and damage levels in four specific classes: "no damage", "minor damage", "major damage", and "destroyed". In order to train the model, proper target masks are needed along with the original images. A target mask is a ground truth annotation image with the same size as the corresponding image but has only one channel. For a pre-disaster image, the mask is a building localization ground truth, which consists of pixel values $0$ for background and $1$ for building. For a post-disaster image, the target mask is damage classification ground truth, pixel values $0$ for background and $\{1,2,3,4\}$ for damage levels. If not provided, we create the necessary ground truth masks from given text-based labels or pre-process them to match the correct format. Table \ref{table_dataset_damage_levels} shows the damage level correspondences for the masks. Figure \ref{fig:sample_xbd_data} shows an example pair of images and masks together used as a single data sample.

\begin{table}[!ht]
    \caption{Damage Levels}
    \label{table_dataset_damage_levels}
    \centering
    \small
    \begin{tabular}{c|c|c}
        \textbf{Damage Level} & \textbf{Name} & \textbf{Pixel Value} \\
        \hline
        L1                    & No Damage      & 1                \\ 
        L2                    & Minor Damage   & 2                \\ 
        L3                    & Major Damage   & 3                \\ 
        L4                    & Destroyed      & 4                \\ 
        \hline
    \end{tabular}
\end{table}

The primary dataset we use is xBD, which is also the sole benchmark used for MambaBDA \cite{ChangeMamba_Chen_2024}. Other than xBD, we utilize Pakistan Flooding, Turkey Earthquake, and Hurricane Ida datasets for training or testing, to further investigate how the suggested methods perform on different domains. The following sections explain statistics and use cases of each dataset.

\subsubsection{xBD}
\label{sec:dataset_xBD}
The xBD dataset \cite{xBD_xView2_gupta2019xbddatasetassessingbuilding} provides the data in four parts: tier 1 and tier 3 splits, which are meant to be used together as the training set, test split, and a holdout split, which we utilized as the validation split. 

While some of the splits come with ready-to-use masks, some of the splits, such as tier 3, did not include target masks necessary for the training. In order to create correct masks, we parsed the .json labels and performed necessary pre-processing operations. Since the labels are given as polygons and coordinates are floating point numbers, depending on the algorithmic method used, there can be small differences between created masks. To ensure the consistency, we also created the target masks for already available splits, such as tier 1, and compare them with the provided ones. We obtained coherent results reaching over $99.8\%$ pixel agreement. 

The dataset is split in 80\%, 10\%, 10\% ratio, containing 9168, 933 and 933 pairs of RGB images in the train, test and validation sets, respectively. The resolution of the images is $1024 \times 1024$ pixels. Table \ref{table_dataset_xbd} displays the damage level distributions of the splits. In the table, the number of images for the damage levels is not meant to sum to the total number of images, as one image may contain more than one damage level or no buildings at all (negative samples). The xBD dataset is used for both training and testing.
\begin{table}[!ht]
    \caption{xBD Dataset Statistics.}
    \label{table_dataset_xbd}
    \centering
    \small
    \begin{tabular}{c|l|r|r|r}
            & \thead{\textbf{Damage} \\ \textbf{Level}} & \textbf{Training} & \textbf{Valid} & \textbf{Test} \\
        \hline
        \multirow{5}{*}{\rotatebox[origin=c]{90}{\thead{ \textbf{Number of} \\ \textbf{Images} }}} 
            & L1     & 4624        & 569             & 570     \\ 
            & L2     & 1785        & 277             & 244     \\ 
            & L3     & 1797        & 320             & 294     \\ 
            & L4     & 2041        & 332             & 288     \\ 
            & Total  & 9168        & 933             & 933     \\ 
        \hline
        \multirow{4}{*}{\rotatebox[origin=c]{90}{\thead{ \textbf{Pixel} \\ \textbf{Ratio} }}} 
            & L1     & 76.35 \%     & 73.29 \%             & 76.67 \%   \\ 
            & L2     & 8.63 \%      & 10.15 \%             & 8.30 \%    \\ 
            & L3     & 8.99 \%      & 10.74 \%             & 9.67 \%    \\ 
            & L4     & 6.02 \%      & 5.82 \%              & 5.37 \%    \\ 
    \end{tabular}
\end{table} 

\subsubsection{Pakistan Flooding}
\label{sec:dataset_pakistanflooding}
Pakistan Flooding is a part of the EBD collection \cite{EBD_Dataset_Wang2024}. The disaster happened in July 2022. The dataset consists of 3540 RGB $512 \times 512$ resolution image pairs. It is not split into subsets so we split it randomly, using 15\% for the validation and test sets and 70\% for the training set. Table \ref{table_dataset_pakistanflooding} shows the dataset's statistics. Pakistan Flooding dataset is used for both training and testing.
\begin{table}[!ht]
    \caption{Pakistan Flooding Dataset Statistics.}
    \label{table_dataset_pakistanflooding}
    \centering
    \small
    \begin{tabular}{c|l|r|r|r}
            & \thead{\textbf{Damage} \\ \textbf{Level}} & \textbf{Training} & \textbf{Valid} & \textbf{Test} \\
        \hline
        \multirow{5}{*}{\rotatebox[origin=c]{90}{\thead{ \textbf{Number of} \\ \textbf{Images} }}} 
            & L1     & 800        & 177             & 174     \\ 
            & L2     & 317        & 70              & 67     \\ 
            & L3     & 155        & 36              & 42     \\ 
            & L4     & 510        & 106             & 119     \\ 
            & Total  & 2478       & 531             & 531     \\ 
        \hline
        \multirow{4}{*}{\rotatebox[origin=c]{90}{\thead{ \textbf{Pixel} \\ \textbf{Ratio} }}} 
            & L1     & 79.58 \%     & 80.87 \%             & 79.86 \%   \\ 
            & L2     & 10.16 \%     & 11.06 \%             & 8.87 \%    \\ 
            & L3     & 2.78 \%      & 2.43 \%             & 3.94 \%    \\ 
            & L4     & 7.48 \%      & 5.64 \%              & 7.33 \%    \\ 
    \end{tabular}
\end{table}

\subsubsection{Turkey Earthquake}
\label{sec:dataset_turkeyearthquake}
Turkey Earthquake is another dataset included in the EBD collection \cite{EBD_Dataset_Wang2024}. It consists of satellite imagery from the devastating earthquakes that occurred in February 2023 in Kahramanmaras, Türkiye. The dataset is smaller compared to the others, also more "no-damage" class skewed, having only 944 image pairs in RGB format in $512 \times 512$ resolution. 
The dataset also includes relatively larger misalignments between pre/post-disaster images compared to other datasets.
This dataset is mainly used as a test dataset, using entirety of the data (combination of all splits) to serve as an unseen test dataset for the models trained on other datasets. However a training experiment also carried out to see how well the model can perform when trained on relatively low amount of misaligned data. Table \ref{table_dataset_turkeyearthquake} shows the statistics for the Turkey Earthquake dataset.
\begin{table}[!ht]
    \caption{Turkey Earthquake Dataset Statistics.}
    \label{table_dataset_turkeyearthquake}
    \centering
    \small
    \begin{tabular}{c|l|r|r|r}
            & \thead{\textbf{Damage} \\ \textbf{Level}} & \textbf{Training} & \textbf{Valid} & \textbf{Test} \\
        \hline
        \multirow{5}{*}{\rotatebox[origin=c]{90}{\thead{ \textbf{Number of} \\ \textbf{Images} }}} 
            & L1     & 489        &  154              & 166     \\ 
            & L2     & 80         &  22               & 19      \\ 
            & L3     & 48         &  13               & 11       \\ 
            & L4     & 102        &  33               & 30      \\ 
            & Total  & 566        &  189              & 189     \\ 
        \hline
        \multirow{4}{*}{\rotatebox[origin=c]{90}{\thead{ \textbf{Pixel} \\ \textbf{Ratio} }}} 
            & L1     & 95.35 \%    & 96.36 \%       & 97.04 \% \\ 
            & L2     & 1.89 \%     & 1.68 \%        & 0.98 \%   \\ 
            & L3     & 1.84 \%     & 0.95 \%        & 0.89 \%   \\ 
            & L4     & 0.92 \%     & 1.01 \%        & 1.08 \%   \\ 
    \end{tabular}
\end{table}

\subsubsection{Hurricane Ida}
Hurricane Ida is a major disaster that took place in Louisiana in 2021. There are several different datasets for this specific incident, including Ida-BD \cite{Ida-BD_lee2022ida} and EBD \cite{EBD_Dataset_Wang2024}. We found that the EBD variant suits our purpose better, since it is a larger alternative. In our experiments we used this dataset for testing only. Thus it serves as a generalization evaluation to an unseen disaster. The test set has 528 RGB images in $512 \times 512$ resolution. The dataset statistics can be seen in Table \ref{table_dataset_hurricaneida}.
\begin{table}[!ht]
    \caption{Hurricane Ida Dataset Statistics.}
    \label{table_dataset_hurricaneida}
    \centering
    \small
    \begin{tabular}{c|l|r}
            & \thead{\textbf{Damage} \\ \textbf{Level}} & \textbf{Test} \\
        \hline
        \multirow{5}{*}{\rotatebox[origin=c]{90}{\thead{ \textbf{Number of} \\ \textbf{Images} }}} 
            & L1       & 438   \\ 
            & L2       & 168   \\ 
            & L3       & 79    \\ 
            & L4       & 35   \\ 
            & Total    & 528   \\ 
        \hline
        \multirow{4}{*}{\rotatebox[origin=c]{90}{\thead{ \textbf{Pixel} \\ \textbf{Ratio} }}} 
            & L1         & 88.67  \%   \\ 
            & L2         & 7.92 \%    \\ 
            & L3         & 3.15 \%    \\ 
            & L4         & 0.26 \%    \\ 
    \end{tabular}
\end{table}

\subsection{Evaluation Metrics}
\label{sec:evaluation_metrics}
Following xView2 Building Damage Assessment challenge \cite{xBD_xView2_gupta2019xbddatasetassessingbuilding} protocol, we report three main metrics:
\begin{itemize}
    \item $F_1^{\text{loc}}$: Denotes the building localization $F_1$ score. 
    \item $F_1^{\text{clf}}$: Denotes the damage classification $F_1$ score. First, per-class $F_1$ scores are computed and then harmonic mean of the four damage scores are used as $F_1^{\text{clf}}$. The classification evaluation is restricted to building pixels, background is ignored for the score calculation. 
    \item $F_1^{\text{oa}}$: Denotes the overall $F_1$ score, which is a weighted mean of the localization and classification scores: $F_1^{\text{oa}} = 0.3 F_1^{\text{loc}} + 0.7F_1^{\text{clf}}$. This is used as the main metric to determine which model performs better.
\end{itemize}
All scores are computed pixel-wise.

\subsection{Training Details}
\label{sec:training_details}
\paragraph{Hyperparameters.} To keep the comparisons consistent, the recommended (or default if not mentioned) values from the ChangeMamba \cite{ChangeMamba_Chen_2024} paper and repository are used as the baseline, with only difference in batch size, 8 vs. 16, due to VRAM limitations. Some notable parameters are: Optimizer is AdamW \cite{adamW_optimizer_loshchilov2019decoupledweightdecayregularization}. Learning Rate is $10^{-4}$. Weight Decay is $5 \times 10^{-3}$. Batch Size is 8. Training data augmentations enabled; random flip, rotation, and $256 \times 256$ crop. Training iterations: $50000$ for xBD, $25000$ for Pakistan Flooding, and $5000$ for Earthquake Turkey datasets.

\paragraph{Environment.} Following hardware and software are used for the training process: RTX 3090 GPU, 24 GB VRAM, CUDA 12.9, Python 3.10.15. Libraries: PyTorch 2.5.0, NumPy 2.2.6.

\subsection{Experiment Groups}
\label{sec:experiment_groups}
\paragraph{Model variants.} As mentioned before, the enhancements are implemented completely in a modular way to better observe each implementation's individual and combined effects. The names and summaries of the evaluated models are as follows:
\begin{itemize}
    \item \textbf{Baseline:} Unmodified MambaBDA model trained using original source code and recommended / default training hyper-parameters.
    \item \textbf{FOCAL:} Baseline model trained with optimized focal loss on damage head. 
    \item \textbf{AGB:} Attention Gates implemented in the building localization head.
    \item \textbf{AGBD:} Attention Gates implemented in both building localization and damage classification heads. 
    \item \textbf{ALIGN:} Custom alignment module integrated into the baseline.
    \item \textbf{Combined Enhancements:} having one or more of the modular enhancements, denoted as FOCAL + ALIGN, FOCAL + AGB, ALIGN + AGB, FOCAL + ALIGN + AGB, and FOCAL + ALIGN + AGBD.
\end{itemize}

Since one of the purposes of this study is to provide improvements with minimal increase in model complexity, we also report the FLOPs and the number of parameters\footnote{Measured with https://github.com/Lyken17/pytorch-OpCounter} of the models for a $512 \times 512$ input size in Table \ref{table_model_complexity_comparison}.
\begin{table}[!ht]
    \caption{Model Complexity Comparison}
    \label{table_model_complexity_comparison}
    \centering
    \small
    \begin{tabular}{l|r|r}
        \textbf{Model / Module} & \textbf{Params (M)} & \textbf{GFLOPs} \\
        \hline
        Baseline    & 87.76     & 195.43  \\ 
        FOCAL       & + 0         & + 0  \\ 
        AGBD        & + 0.10     & + 0.71  \\ 
        AGB         & + 0.05     & + 0.36  \\ 
        ALIGN       & + 0.63     & + 0.65  \\ 
    \end{tabular}
\end{table}

\paragraph{Evaluation categories.} The evaluations are structured under two categories. In-domain experiments train and test the models within the same dataset using its own test split. This setting measures how well a model performs in a known environment. xBD, Pakistan Flooding, and Turkey Earthquake datasets are used for in-domain tests. In cross-dataset experiments the models are tested on unseen datasets. For example the models trained on xBD or Pakistan Flooding datasets are tested on each other's test splits, or they are both tested on Turkey Earthquake and Hurricane Ida datasets. This setting measures how well a model generalizes. Each experiment is conducted with all of the model variants explained above. 


\section{\uppercase{Results}}
\label{sec:results}
In this section, we analyze the results of in-domain and cross-dataset tests, and discuss the models' performance. 

\subsection{In-domain Tests}
\label{sec:indomain_tests}
These experiments are carried out on xBD, Pakistan Flooding and Turkey Earthquake datasets and detailed results can be seen in the tables \ref{table_results_indomain_xbd}, \ref{table_results_indomain_pakistanflooding} and \ref{table_results_indomain_turkeyearthquake}. In all of the tables, the best overall $F_1$ score is written in \textcolor{blue}{\textbf{blue}}, and the next two best results are highlighted in \textbf{bold}. All reported increases in the score are \textbf{absolute} changes in percentage, not relative to the baseline score. Since the models trained on xBD and Pakistan Flooding datasets are also used in cross-dataset tests, they will be referred as Group 1 and Group 2, respectively.

On xBD, the best overall score is achieved by the \textit{FOCAL + ALIGN + AGB} model, improving the baseline score by $1.04\%$. Most of the modular enhancements actually improve the baseline model except for the \textit{AGBD} model. It can be seen that \textit{FOCAL} improves the classification scores, mostly "L3-Major". \textit{ALIGN} slightly contributes to localization, and \textit{AGB} provides gain on both localization and classification. 

\begin{table*}[!ht]
    \caption{In-Domain Evaluation Results on xBD Dataset (Group 1)}
    \label{table_results_indomain_xbd}
    \centering
    \small
    \begin{tabular}{l|cc:c|cccc}
        \textbf{Model} & $F_1^{\text{loc}}$ & $F_1^{\text{clf}}$ & $F_1^{\text{oa}}$ 
        & $F_1^{\text{L1}}$ & $F_1^{\text{L2}}$ & $F_1^{\text{L3}}$ & $F_1^{\text{L4}}$ \\
        \hline
        Baseline                & 84.86 & 75.23 & 78.12 & 95.23 & 57.36 & 73.02 & 86.66 \\
        \hline
        AGBD                    & 84.48 & 74.98 & 77.83 & 95.13 & 55.55 & 75.17 & 86.74 \\
        AGB                     & 85.42 & 76.29 & \textbf{79.03} & 94.75 & 59.32 & 74.71 & 85.92 \\
        ALIGN                   & 85.14 & 75.76 & 78.57 & 95.45 & 57.27 & 75.19 & 86.48 \\
        FOCAL                   & 84.84 & 75.77 & 78.49 & 94.56 & 57.97 & 74.30 & 86.90 \\
        \hline
        FOCAL + ALIGN           & 84.68 & 76.12 & 78.69 & 95.38 & 58.41 & 75.10 & 85.99 \\
        FOCAL + AGB             & 85.09 & 76.11 & 78.81 & 94.20 & 60.62 & 72.31 & 86.11 \\
        ALIGN + AGB             & 84.98 & 76.54 & \textbf{79.07} & 94.93 & 59.95 & 74.87 & 85.57 \\
        FOCAL + ALIGN + AGBD    & 84.70 & 76.39 & 78.89 & 95.05 & 58.75 & 75.13 & 86.89 \\
        FOCAL + ALIGN + AGB     & 85.26 & 76.55 & \textcolor{blue}{\textbf{79.16}} & 95.26 & 60.36 & 73.84 & 85.85 \\
    \end{tabular}
\end{table*}

\begin{figure}
    \centering
    \includegraphics[width=\linewidth]{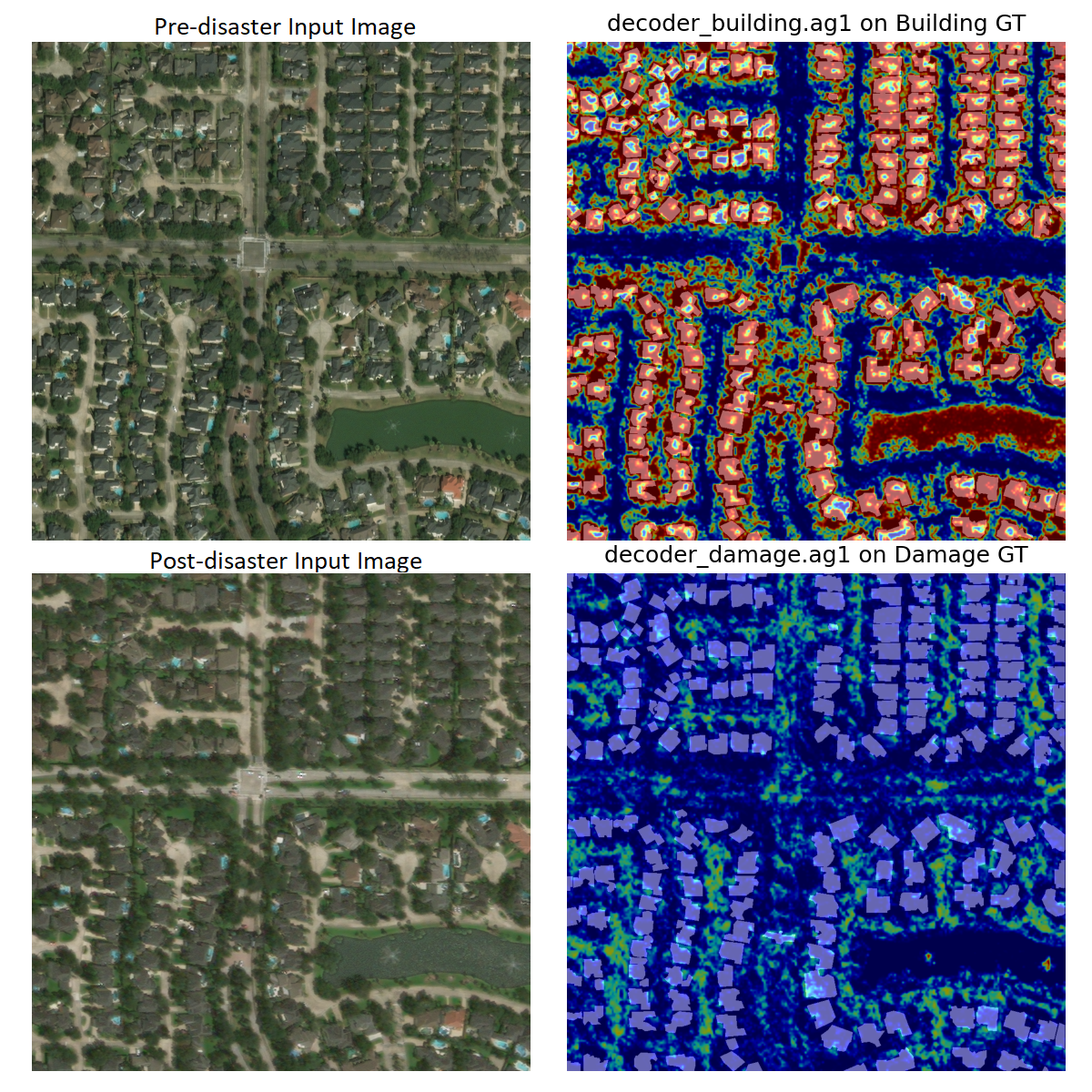}
    \caption{Example attention gate outputs. Heatmaps are overlayed on ground truth masks. Attention increases from blue to green to red. The top right heatmap belongs to AGB, the AGs in building decoder, and the bottom right heatmap belongs to AGD, the AGs in damage decoder. AGB mostly focuses on building borders improving performance, while AGD mostly focuses on spaces between the buildings, which does not increase the scores.}
    \label{fig:attention_gate_outputs}
\end{figure}

\textit{AGBD} on the other hand does not improve the baseline model. When the model's intermediate outputs are reviewed, it can be seen that the attention gates in the damage decoder does not learn properly and either does not guide the model, being noneffective, or makes the model focus on incorrect parts, degrading the scores. Figure \ref{fig:attention_gate_outputs} shows an example output of the attention maps. Throughout the experiments, we saw that damage attention gates sometimes learn correctly, increasing the model's performance, but sometimes learn in a way that harms performance, which makes it unstable. This causes the damage head AGs to be an unreliable module. It can be seen from the Turkey Earthquake tests in Table \ref{table_results_indomain_turkeyearthquake} that \textit{AGBD} improves the classification results of all damage classes, whereas In Pakistan Flooding, it only helps with the "L4-Destroyed" class, and decreases other classification results. In contrast, \textit{FOCAL + AGB} models consistently improve the baseline performance in all datasets. Focal loss primarily increases the minority class scores, focusing on the imbalance problem. \textit{AGB} reduces false positives and increases true positives, improving localization and, this way, also indirectly classification scores.

In the Pakistan Flooding dataset, the best performance is achieved by \textit{FOCAL + AGB}, which provides a $0.82\%$ increase from the baseline, with notable improvement on "L3-Major" class ($3.06\%$ increase). In the Turkey Earthquake dataset, the best performing model is \textit{AGB}. \textit{ALIGN} in this case increases the localization scores relatively more than in the other datasets. This stems from Turkey Earthquake dataset having more misalignments compared to the other datasets, proving the benefits of the module.

\begin{table*}[!ht]
    \caption{In-Domain Evaluation Results on Pakistan Flooding Dataset (Group 2)}
    \label{table_results_indomain_pakistanflooding}
    \centering
    \small
    \begin{tabular}{l|cc:c|cccc}
        \textbf{Model} & $F_1^{\text{loc}}$ & $F_1^{\text{clf}}$ & $F_1^{\text{oa}}$ 
        & $F_1^{\text{L1}}$ & $F_1^{\text{L2}}$ & $F_1^{\text{L3}}$ & $F_1^{\text{L4}}$ \\
        \hline
        Baseline                & 87.67 & 77.44 & 80.51 & 96.66 & 74.03 & 69.24 & 74.87 \\
        \hline
        AGBD                    & 87.81 & 76.99 & 80.24 & 96.59 & 73.86 & 67.27 & 75.77 \\
        AGB                     & 87.75 & 76.46 & 79.85 & 97.04 & 78.07 & 63.03 & 74.98 \\
        ALIGN                   & 87.58 & 78.01 & \textbf{80.88} & 96.49 & 74.20 & 68.75 & 77.58 \\
        FOCAL                   & 88.15 & 77.63 & \textbf{80.78} & 96.71 & 74.86 & 69.02 & 74.95 \\
        \hline
        FOCAL + ALIGN           & 88.43 & 77.19 & 80.56 & 96.77 & 73.64 & 66.66 & 77.50 \\
        FOCAL + AGB             & 88.15 & 78.40 & \textcolor{blue}{\textbf{81.33}} & 96.51 & 73.60 & 72.30 & 75.54 \\
        ALIGN + AGB             & 87.92 & 76.59 & 79.98 & 97.14 & 74.52 & 66.05 & 74.77 \\
        FOCAL + ALIGN + AGBD    & 88.34 & 76.95 & 80.37 & 96.03 & 75.22 & 65.70 & 76.61 \\
        FOCAL + ALIGN + AGB     & 88.26 & 77.43 & 80.68 & 96.68 & 76.47 & 68.51 & 73.31 \\
    \end{tabular}
\end{table*}

\begin{table*}[!ht]
    \caption{In-Domain Evaluation Results on Turkey Earthquake Dataset}
    \label{table_results_indomain_turkeyearthquake}
    \centering
    \small
    \begin{tabular}{l|cc:c|cccc}
        \textbf{Model} & $F_1^{\text{loc}}$ & $F_1^{\text{clf}}$ & $F_1^{\text{oa}}$ 
        & $F_1^{\text{L1}}$ & $F_1^{\text{L2}}$ & $F_1^{\text{L3}}$ & $F_1^{\text{L4}}$ \\
        \hline
        Baseline              & 90.69 & 29.97 & 48.18 & 98.43 & 20.13 & 17.06 & 66.63 \\
        \hline
        AGBD                  & 90.75 & 35.00 & 51.72 & 98.96 & 22.44 & 22.09 & 69.63 \\
        AGB                   & 91.31 & 36.86 & \textcolor{blue}{\textbf{53.20}} & 98.94 & 27.14 & 21.27 & 68.76 \\
        ALIGN                 & 91.28 & 29.06 & 47.73 & 98.98 & 15.75 & 20.05 & 70.61 \\
        FOCAL                 & 91.19 & 30.33 & 48.59 & 98.92 & 20.29 & 17.17 & 70.28 \\
        \hline
        FOCAL + ALIGN         & 91.22 & 29.51 & 48.02 & 98.92 & 15.14 & 22.29 & 68.77 \\
        FOCAL + AGB           & 90.69 & 35.14 & \textbf{51.81} & 98.72 & 33.05 & 16.76 & 72.51 \\
        ALIGN + AGB           & 91.24 & 27.53 & 46.64 & 98.92 & 20.15 & 13.71 & 79.21 \\
        FOCAL + ALIGN + AGBD  & 90.87 & 36.20 & \textbf{52.60} & 98.46 & 30.51 & 20.19 & 55.47 \\
        FOCAL + ALIGN + AGB   & 91.16 & 28.85 & 47.54 & 97.78 & 17.84 & 17.60 & 64.25 \\
    \end{tabular}
\end{table*}

As a result, the investigated modular modifications are mostly capable of improving the baseline performance in in-domain tasks. The overall performance is improved by $0.8\%$ to $5\%$ on different datasets.


\subsection{Cross-dataset Tests}
\label{sec:crossdataset_tests}
Cross-dataset tests are for generalization capability measurements. When tested on unseen datasets, the models' performances drop significantly. Our modular enhancements mitigate this problem partially. The tables \ref{table_results_crossdomain_xbd_on_pakistanflooding}, \ref{table_results_crossdomain_xbd_on_hurricaneida}, and \ref{table_results_crossdomain_xbd_on_turkeyearthquake} show the performances of the models trained on xBD (Group 1) and tested  on the Pakistan Flooding, Hurricane Ida, and Turkey Earthquake datasets, respectively. Similarly, the tables \ref{table_results_crossdomain_pakistanflooding_on_xbd}, \ref{table_results_crossdomain_pakistanflooding_on_hurricaneida}, and \ref{table_results_crossdomain_pakistanflooding_on_turkeyearthquake} display the performance of the models trained on Pakistan Flooding (Group 2) and tested on the xBD, Hurricane Ida, and Turkey Earthquake datasets, respectively.

Group 1 performs better than the Group 2 in general, which is expected since xBD is larger and consists of more variable disasters. When evaluated on Pakistan Flooding dataset, \textit{FOCAL + AGB} combination achieves an overall score of $56.60\%$ when the baseline is only $29.56\%$, demonstrating a significant improvement. On Hurricane Ida, \textit{AGB} and \textit{ALIGN} provide notable gains, with \textit{ALIGN + AGB} combination performing the best with $5.86\%$ increase. On the Turkey Earthquake dataset, while both the general scores and improvements are smaller, \textit{FOCAL + AGB} again represents the most effective configuration as it increases the score by $1.88\%$.

\begin{table*}[!ht]
    \caption{Cross-Dataset Evaluation of xBD-Trained Models (Group 1) on Pakistan Flooding Dataset}
    \label{table_results_crossdomain_xbd_on_pakistanflooding}
    \centering
    \small
    \begin{tabular}{l|cc:c|cccc}
        \textbf{Model} & $F_1^{\text{loc}}$ & $F_1^{\text{clf}}$ & $F_1^{\text{oa}}$ 
        & $F_1^{\text{L1}}$ & $F_1^{\text{L2}}$ & $F_1^{\text{L3}}$ & $F_1^{\text{L4}}$ \\
        \hline
        Baseline              & 60.27 & 16.40 & 29.56 & 91.30 &  5.42 & 41.44 & 41.01 \\
        \hline
        AGBD                  & 63.39 & 24.36 & 36.07 & 91.46 &  9.41 & 44.74 & 40.55 \\
        AGB                   & 63.93 & 34.94 & 43.63 & 93.67 & 17.32 & 45.60 & 41.41 \\
        ALIGN                 & 53.75 & 38.76 & 43.26 & 90.83 & 23.25 & 47.16 & 35.76 \\
        FOCAL                 & 61.56 & 43.00 & \textbf{48.57} & 91.76 & 33.42 & 44.25 & 33.78 \\
        \hline
        FOCAL + ALIGN         & 57.35 & 37.15 & 43.21 & 85.34 & 23.79 & 44.43 & 31.84 \\
        FOCAL + AGB           & 66.73 & 52.27 & \textcolor{blue}{\textbf{56.60}} & 94.62 & 54.29 & 44.25 & 40.09 \\
        ALIGN + AGB           & 63.41 & 41.30 & \textbf{47.93} & 94.18 & 46.62 & 43.78 & 23.84 \\
        FOCAL + ALIGN + AGBD  & 55.32 & 42.82 & 46.57 & 94.86 & 36.42 & 43.58 & 30.81 \\
        FOCAL + ALIGN + AGB   & 61.31 & 26.72 & 37.10 & 94.28 & 44.78 & 39.13 & 10.96 \\ 
    \end{tabular}
\end{table*}

\begin{table*}[!ht]
    \caption{Cross-Dataset Evaluation of xBD-Trained Models (Group 1) on Hurricane Ida Dataset}
    \label{table_results_crossdomain_xbd_on_hurricaneida}
    \centering
    \small
    \begin{tabular}{l|cc:c|cccc}
        \textbf{Model} & $F_1^{\text{loc}}$ & $F_1^{\text{clf}}$ & $F_1^{\text{oa}}$ 
        & $F_1^{\text{L1}}$ & $F_1^{\text{L2}}$ & $F_1^{\text{L3}}$ & $F_1^{\text{L4}}$ \\
        \hline
        Baseline              & 88.95 & 19.98 & 40.67 & 90.20 & 21.54 & 23.75 &  9.95 \\
        \hline
        AGBD                  & 86.27 & 14.85 & 36.28 & 93.08 & 24.81 &  5.74 & 22.64 \\
        AGB                   & 88.02 & 24.79 & \textbf{43.76} & 89.27 & 29.32 & 20.11 & 15.07 \\
        ALIGN                 & 89.18 & 24.99 & \textbf{44.24} & 92.24 & 31.58 & 35.64 & 11.17 \\
        FOCAL                 & 88.90 & 16.37 & 38.13 & 88.87 & 26.96 & 51.75 &  5.66 \\
        \hline
        FOCAL + ALIGN         & 87.84 & 21.37 & 41.31 & 91.77 & 34.24 & 37.44 &  8.31 \\
        FOCAL + AGB           & 88.34 & 18.27 & 39.29 & 89.57 & 28.09 & 21.78 &  7.92 \\
        ALIGN + AGB           & 87.26 & 29.08 & \textcolor{blue}{\textbf{46.53}} & 93.65 & 18.88 & 47.81 & 18.86 \\
        FOCAL + ALIGN + AGBD  & 85.16 &  8.11 & 31.22 & 93.29 & 13.00 &  3.45 &  8.62 \\
        FOCAL + ALIGN + AGB   & 88.21 & 23.77 & 43.10 & 90.94 & 22.81 & 52.54 & 10.59 \\
    \end{tabular}
\end{table*}

\begin{table*}[!ht]
    \caption{Cross-Dataset Evaluation of xBD-Trained Models (Group 1) on Turkey Earthquake Dataset}
    \label{table_results_crossdomain_xbd_on_turkeyearthquake}
    \centering
    \small
    \begin{tabular}{l|cc:c|cccc}
        \textbf{Model} & $F_1^{\text{loc}}$ & $F_1^{\text{clf}}$ & $F_1^{\text{oa}}$ 
        & $F_1^{\text{L1}}$ & $F_1^{\text{L2}}$ & $F_1^{\text{L3}}$ & $F_1^{\text{L4}}$ \\
        \hline
        Baseline              & 84.04 &  0.02 & 25.23 & 97.08 &  0.00 &  2.52 & 18.32 \\
        \hline
        AGBD                  & 84.08 &  0.00 & 25.22 & 97.89 &  0.00 &  1.09 & 18.50 \\
        AGB                   & 84.36 &  0.72 & \textbf{25.81} & 97.76 &  0.20 &  2.23 & 28.40 \\
        ALIGN                 & 82.95 &  0.16 & 24.99 & 96.43 &  0.04 &  1.63 & 17.73 \\
        FOCAL                 & 84.63 &  0.08 & 25.45 & 96.83 &  0.02 &  1.22 & 25.61 \\
        \hline
        FOCAL + ALIGN         & 82.90 &  0.00 & 24.87 & 95.10 &  0.00 & 11.25 & 15.36 \\
        FOCAL + AGB           & 85.52 &  2.07 & \textcolor{blue}{\textbf{27.11}} & 97.80 &  1.64 &  0.84 &  8.73 \\
        ALIGN + AGB           & 83.56 &  2.10 & \textbf{26.54} & 97.43 &  0.59 &  7.71 & 11.93 \\
        FOCAL + ALIGN + AGBD  & 80.71 &  2.03 & 25.64 & 97.79 &  0.94 &  1.17 & 26.32 \\
        FOCAL + ALIGN + AGB   & 84.04 &  0.00 & 25.21 & 97.87 &  0.00 &  0.32 & 18.13 \\
    \end{tabular}
\end{table*}


In contrast to Group 1, Group 2 has limited generalization capability as Pakistan Flooding is not as diverse as xBD. When Group 2 is evaluated on xBD, the best performing models are \textit{FOCAL + AGB} and \textit{FOCAL + ALIGN + AGB}. It can be seen that focal loss increases the classification scores of the classes the baseline model confuses most. Although \textit{ALIGN} and \textit{AGB} individually performs poorly, when combined with \textit{FOCAL}, they also contribute to the localization score improvement. This combination of \textit{FOCAL + ALIGN + AGB} increases the overall $F_1$ by $9.56\%$, which indicates better generalization. On Hurricane Ida dataset, \textit{ALIGN} again performs poorly, while the best performing models are \textit{FOCAL + AGB} and \textit{\textit{FOCAL + ALIGN + AGB}} increasing the score by $3.00\%$. Finally on the Turkey Earthquake dataset, although the improvements are smaller,  the \textit{FOCAL + ALIGN + AGBD} model manages to improve the overall score by $1.08\%$ by improving the classification scores of hardest classes, "L2-Minor" and "L3-Major".

\begin{table*}[!ht]
    \caption{Cross-Dataset Evaluation of Pakistan Flooding-Trained Models (Group 2) on xBD Dataset}
    \label{table_results_crossdomain_pakistanflooding_on_xbd}
    \centering
    \small
    \begin{tabular}{l|cc:c|cccc}
        \textbf{Model} & $F_1^{\text{loc}}$ & $F_1^{\text{clf}}$ & $F_1^{\text{oa}}$ 
        & $F_1^{\text{L1}}$ & $F_1^{\text{L2}}$ & $F_1^{\text{L3}}$ & $F_1^{\text{L4}}$ \\
        \hline
        Baseline              & 40.13 &  0.51 & 12.39 & 86.58 &  0.13 & 12.63 & 12.08 \\
        \hline
        AGBD                  & 33.59 & 12.22 & 18.63 & 86.37 &  4.75 & 34.17 & 13.18 \\
        AGB                   & 31.64 &  5.95 & 13.65 & 77.46 &  1.79 & 23.24 & 17.36 \\
        ALIGN                 & 14.10 &  2.83 &  6.21 & 86.78 &  0.80 & 16.27 & 10.78 \\
        FOCAL                 & 33.64 &  6.54 & 14.67 & 83.65 &  2.20 & 27.44 &  9.19 \\
        \hline
        FOCAL + ALIGN         & 48.13 &  6.03 & \textbf{18.66} & 80.55 &  1.83 & 24.91 & 15.87 \\
        FOCAL + AGB           & 48.10 &  8.60 & \textbf{20.45} & 86.32 &  2.97 & 34.90 & 11.38 \\
        ALIGN + AGB           & 44.67 &  0.64 & 13.84 & 86.71 &  0.16 & 27.60 & 10.52 \\
        FOCAL + ALIGN + AGBD  & 35.22 &  9.42 & 17.16 & 79.60 &  3.07 & 30.20 & 18.74 \\
        FOCAL + ALIGN + AGB   & 47.11 & 11.14 & \textcolor{blue}{\textbf{21.93}} & 85.65 &  4.30 & 22.57 & 14.22 \\
    \end{tabular}
\end{table*}

\begin{table*}[!ht]
    \caption{Cross-Dataset Evaluation of Pakistan Flooding-Trained Models (Group 2) on Hurricane Ida Dataset}
    \label{table_results_crossdomain_pakistanflooding_on_hurricaneida}
    \centering
    \small
    \begin{tabular}{l|cc:c|cccc}
        \textbf{Model} & $F_1^{\text{loc}}$ & $F_1^{\text{clf}}$ & $F_1^{\text{oa}}$ 
        & $F_1^{\text{L1}}$ & $F_1^{\text{L2}}$ & $F_1^{\text{L3}}$ & $F_1^{\text{L4}}$ \\
        \hline
        Baseline              & 47.60 &  0.36 & 14.53 & 93.35 &  0.11 &  0.62 &  2.88 \\
        \hline
        AGBD                  & 46.21 &  2.62 & 15.69 & 91.63 &  1.50 & 16.95 &  1.26 \\
        AGB                   & 39.55 &  3.45 & 14.28 & 87.16 &  1.53 & 11.61 &  2.46 \\
        ALIGN                 & 29.79 &  0.00 &  8.94 & 93.70 &  0.79 &  0.00 &  4.56 \\
        FOCAL                 & 42.89 &  3.59 & 15.38 & 90.46 &  1.85 &  7.74 &  2.30 \\
        \hline
        FOCAL + ALIGN         & 51.78 &  0.48 & 15.87 & 89.61 &  0.13 & 19.49 &  3.34 \\
        FOCAL + AGB           & 53.94 &  1.89 & \textbf{17.50} & 91.91 &  0.55 & 11.24 &  4.76 \\
        ALIGN + AGB           & 47.34 &  0.27 & 14.39 & 93.14 &  0.07 &  5.37 &  2.16 \\
        FOCAL + ALIGN + AGBD  & 48.29 &  3.93 & \textbf{17.24} & 86.61 &  1.90 & 14.77 &  2.42 \\
        FOCAL + ALIGN + AGB   & 48.46 &  4.28 & \textcolor{blue}{\textbf{17.53}} & 91.05 &  1.98 &  7.55 &  3.49 \\
    \end{tabular}
\end{table*}

\begin{table*}[!ht]
    \caption{Cross-Dataset Evaluation of Pakistan Flooding-Trained Models (Group 2) on Turkey Earthquake Dataset}
    \label{table_results_crossdomain_pakistanflooding_on_turkeyearthquake}
    \centering
    \small
    \begin{tabular}{l|cc:c|cccc}
        \textbf{Model} & $F_1^{\text{loc}}$ & $F_1^{\text{clf}}$ & $F_1^{\text{oa}}$ 
        & $F_1^{\text{L1}}$ & $F_1^{\text{L2}}$ & $F_1^{\text{L3}}$ & $F_1^{\text{L4}}$ \\
        \hline
        Baseline              & 64.93 &  0.00 & 19.48 & 96.95 &  0.57 &  0.00 &  4.47 \\
        \hline
        AGBD                  & 55.88 &  0.00 & 16.77 & 96.99 &  0.11 &  0.00 &  2.28 \\
        AGB                   & 54.74 &  0.24 & 16.59 & 92.12 &  1.43 &  0.06 &  7.75 \\
        ALIGN                 & 52.87 &  0.00 & 15.86 & 97.26 &  0.77 &  0.00 &  1.96 \\
        FOCAL                 & 62.27 &  1.18 & 19.50 & 94.86 &  1.69 &  0.38 &  6.67 \\
        \hline
        FOCAL + ALIGN         & 64.17 &  0.46 & \textbf{19.57} & 92.21 &  0.20 &  0.28 &  2.94 \\
        FOCAL + AGB           & 64.52 &  0.00 & 19.36 & 96.94 &  0.62 &  0.00 &  0.79 \\
        ALIGN + AGB           & 61.04 &  1.17 & 19.13 & 96.46 &  0.66 &  0.80 &  1.53 \\
        FOCAL + ALIGN + AGBD  & 64.13 &  1.89 & \textcolor{blue}{\textbf{20.56}} & 94.70 &  1.96 &  0.74 &  4.25 \\
        FOCAL + ALIGN + AGB   & 63.67 &  1.62 & \textbf{20.24} & 93.70 &  1.67 &  0.60 &  5.48 \\
    \end{tabular}
\end{table*}

\subsection{Discussion}
It can be seen from the experiment results that the proposed modular enhancements indeed improve performance. 
For in-domain tests, the improvements are modest but consistent, providing $0.8\%$ to $5\%$ performance gain overall. The best resulting models are mostly \textit{FOCAL} and \textit{AGB}, while \textit{ALIGN} also contributes depending on the misalignments. For the cross-dataset tests, the improvements are even more successful. The baselines did not perform well on unseen datasets, and our models can improve the scores by large amounts up to $27\%$ depending on the setup.

In general, focal loss usually improves the classification scores of the hardest classes, i.e., the classes that the baseline model has the lowest scores on. AGB module also consistently improves scores, mostly both localization, by guiding the model to focus on correct spots, and classification, sometimes indirectly by increasing true positive building regions thus increasing new correct classifications. These are helpful in both in-domain and new domain scenarios. On the other hand, alignment module have very limited transferability, especially when trained on a non-diverse dataset, but performs well in-domain.


\section{\uppercase{Conclusions}}
\label{sec:conclusion}

In this work, we investigate three ways to improve the state-of-the-art building damage assessment architecture MambaBDA. We demonstrate that the model's performance can be further increased with design improvements for the problem at hand. The explored enhancements are: focal loss for class imbalance, attention gates for better feature selection, and alignment module for handling spatial misalignments. These methods are intentionally designed and implemented in a modular way for simple integration, without depending on the backbone. They are also computationally inexpensive. 

In order to evaluate the methods, we conduct detailed tests on five different datasets. We perform both in-domain and cross-dataset evaluations using all individual and combined implementations of our modular enhancements along with the baseline model. The results show that the methods provide consistent improvements in both experimental settings, especially in cross-dataset experiments with up to $27\%$ $F_1$ score increase. 





\bibliographystyle{apalike}
{\small
\bibliography{main}}




\end{document}